\documentclass[runningheads]{llncs}
\usepackage{graphicx}
\usepackage{wrapfig}
\usepackage{caption}
\usepackage{amsmath}
\usepackage{nccmath}
\usepackage{amsfonts}
\usepackage{amssymb}

\begin{document}
\title{Quantization Aware Training, ERNIE and Kurtosis Regularizer: \\ an empirical study}
\titlerunning{Quantization Aware Training, ERNIE and Kurtosis Regularizer}
%
\author{Andrea Zanetti} 
%
%
\institute{
Intel Technology Poland, Gdansk, Poland \\
\email{{andrea.zanetti@intel.com}}
}
\maketitle              
\begin{abstract}
Pre-trained language models like ERNIE or BERT \cite{ERNIE} are currently used in many applications. These models come with a set of pre-trained weights typically obtained in unsupervised/self-supervised modality on a huge amount of data. After that, they are fine-tuned on a specific task. Applications then use these models for inference, and often some additional constraints apply, like low power-budget or low latency between input and output.
The main avenue to meet these additional requirements for the inference settings, is to use low precision computation (e.g. INT8 rather than FP32), but this comes with the cost of deteriorating the functional performance (e.g. accuracy) of the model. Some approaches have been developed to tackle the problem and go beyond the limitations of the PTO (Post-Training Quantization), more specifically the QAT (Quantization Aware Training, see \cite{QAT}) is a procedure that “interferes” with the training process in order to make it affected (or simply ‘disturbed’) by the quantization phase during the training itself. Besides QAT, recently Intel-Habana Labs have proposed an additional and more direct way to make the training results more robust to subsequent quantization which uses a regularizer, therefore changing the loss function that drives the training procedure. But their proposal does not work “out of the box” for pre-trained models like ERNIE, for example. In this short paper we show why this is not happening (for the ERNIE case) and we propose a very basic way to deal with it, sharing as well some initial results (increase in final INT8 accuracy) that might be of interest to practitioners willing to use ERNIE in their applications, in low precision regime.
\end{abstract}

\section{Introduction: Quantization Aware Training - what it is and why it is useful }
Quantization Aware Training (QAT) is a framework for training neural networks that modifies the training procedure inserting (in the forward pass) fake-quantize operations that simulate the actual quantization of the computations (and of data used for it), typically leaving the backward pass untouched. The overall goal of this technique is to make the final numeric format -that will be used at inference time- affect the training procedure, in such a way to make the final solution found by the learning algorithm more compatible and thus better performant with the target low precision numeric format.

Among the entities that are quantized, most often there are some -if not all- the parameters of the network. As mentioned above, in doing so the training procedure is pushed to converge towards solutions (aka set of parameters) that are likely to be more robust -in terms of degradation of performance- with respect the quantization of the model itself. However, there are very many possible ways to quantize a model. Typically, the initial numerical format is FP32, which we assume here to be the format used for training, whereas the destination numeric format, that we want to use at inference time, could be FP16 or more commonly, INT8.
For simplicity here we focus on a quantization strategy that is simple but at the same time widely used, and therefore of practical importance; we will steer our attention on the \textbf{symmetric uniform FP32 - to - INT8 quantization} case, in which a tensor of FP32 numbers is represented by a single FP32 numbers (scale) and an INT8 tensor of the same size of the original FP32 tensor.

\textbf{But how this mapping works?} There are many techniques for choosing the most appropriate representation of a FP32 tensor using a FP32 scale and a corresponding tensor of INT8 numbers. For example, we can choose the scale using the ABS\_MAX approach as depicted in fig.\ref{fig:MAXABSexample} (but the reader must be warned that other approaches are possible): 

\begin{figure}
\vspace{-1.8\baselineskip}
\includegraphics[width=1.0\linewidth, height=4cm]{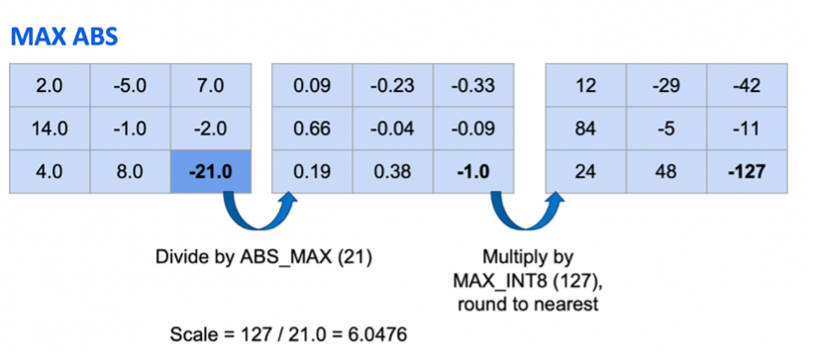} 
\caption{\label{fig:MAXABSexample} Example of the MAX\_ABS approach to choose the scale for the conversion of a tensor from FP32 into INT8.}
\vspace{-0.8\baselineskip}
\end{figure}

\begin{figure}
\vspace{-1.8\baselineskip}
\includegraphics[width=1.0\linewidth, height=4cm]{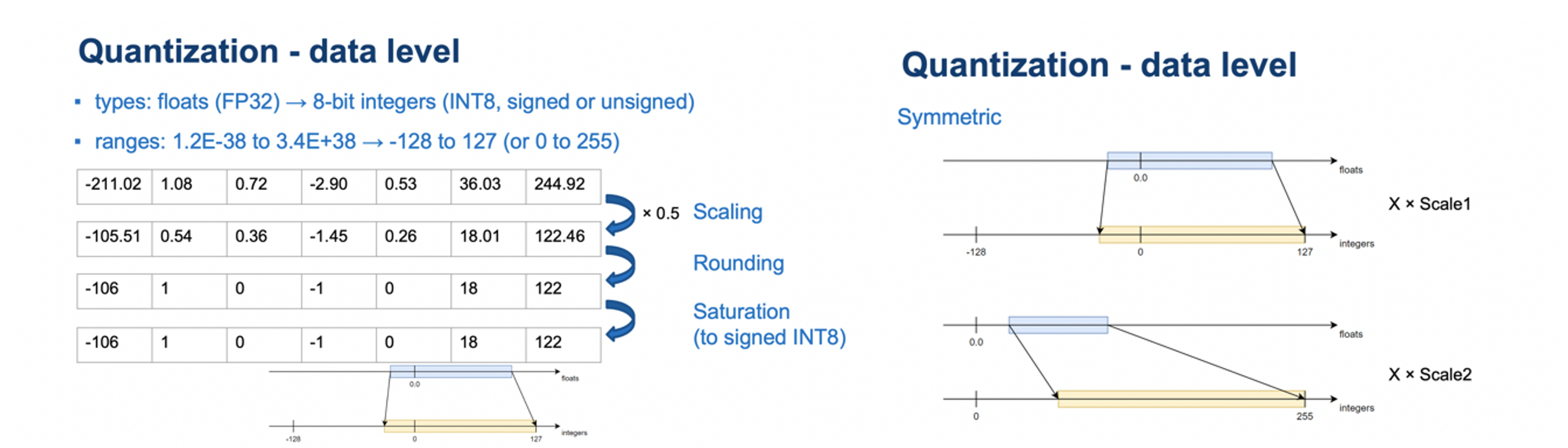} 
\caption{\label{fig:SymQuant} Symmetric quantization: visualization of the overall process.}
\vspace{-0.8\baselineskip}
\end{figure}

In a nutshell, it can be said the QAT procedure makes the training procedure aware of the quantization step and it does that in an implicit way due to the fact that the quantization step is experienced by the network during training.

However, we may want to add an explicit push for the training towards specific characteristics of the tensors that are useful for the purpose of using at interference time our model with a reduce precision. But what characteristics for our tensors should we enforce in order to make the quantization of a model the least harmful possible? And how practically implement this enforcement? 

\section{A very smart idea from our colleagues in Intel-Habana Labs}
Intel-Habana Labs have recently published a paper \cite{Habana_Kure} which considers the problem of steering the training procedure in such a way that the training itself is guided to finding solutions (that is, a set of parameters in parameter space) such that the subsequent quantization of the model is minimally harmful. 
Their paper considers the case of uniform symmetrical quantization and shows that in such settings, among all possible distributions, the uniform distribution is the one that yields the minimal quantization sensitivity (theorem 4 in the paper).
The technique they devised to guide the training is to add a regularizer to the loss function which is computed using the distribution of all the weight tensors in the model at hand.
More precisely, they enforce the preference of the training procedure for solutions which have uniformly distributed weight tensors adding a regularizer that is obtained from the sum of the Kurtosis of each weight tensors.
As a reminder, given a set of n numbers (for example, a tensor T in the Deep Learning parlance), its Kurtosis is calculated as follows, with the operator '$E$' being the probabilistic expectation:

\begin{eqnarray}
K_x &= \frac{E\left[\left( x - \mu \right)^4 \right]}{\left(E\left[\left( x - \mu \right)^2 \right]\right)^{2}}
&=  E\left[\left( x - \mu \right)^4 \right] \left( E\left[\left( x - \mu \right)^2 \right] \right)^{-2}
\end{eqnarray}

Even though the relation between distributions and Kurtosis values is not a bijection, it turns out to be practically useful to take the value of Kurtosis of the uniform distribution as a proxy of uniformity; that is, forcing the training procedure to choose weight tensors which have the same Kurtosis value as the uniform distribution is likely to push those tensors towards a more uniform distribution.
This fact, that in theory is not guaranteed, in practice is well established especially for the models trained from scratch, where the initial parameters distributions are typically Normal-like and surely well verified for the models considered in the Intel-Habana Labs paper, which are mostly based on convolutional architectures. 
Their paper \cite{Habana_Kure} concludes with the statement: 

\textit{“This work focuses on weights but can also be used for activations. 
KURE (the regularization based on Kurtosis) can be extended to 
other domains such as recommendation systems and NLP models”}

So, along with a QAT procedure, it is now also possible to deal with Kurtosis regularizer to explicitly “tell” the training procedure to prefer more uniformly distributed tensors.
But will the Kurtosis regularized work out of the box also for pre-trained models that need fine-tuning before being eventually quantized and deployed on the target device?
In addition, when considering both strategies (QAT and Kurtosis regularizer) is better to use them at the same time or pipeline them?
The ambition of this short paper is not to provide precise answer to these questions, but to share the results of some experiments we have done in the context of QAT with an important pre-trained model such as ERNIE, and the insertion of KURE, that is name given to the Kurtosis regularizer, in the process

\section{A special case of practical importance:  applying kurtosis to pretrained models – the ERNIE case }
Here we present a case of application of Kurtosis as regularizer that is not considered in the reference paper \cite{Habana_Kure}, but which is of practical interest. \\
The problem we consider is the following: pretrained models such as BERT or ERNIE come with a lot of knowledge encoded in their parameters, thanks to the huge amount of data that has been used for the pretraining itself, and the smart pretraining techniques used. Typically, in order to fully leverage their transfer learning potential, intermediate users of such models add one or few layers on top of these models and do some fine tuning of the whole Neural Network so obtained in order to get a final model for their task.
The final model then might be deployed on devices where the power budget is constrained or where the memory footprint is critical, or even where the latency should be minimized. 
And here is where QAT comes into play in order to get a low-precision version of the ERNIE fine-tuned model that can be deployed on the target device.
Clearly the hope is to retain as much as possible the performance of the FP32-original model into the final INT8 model. For all these cases, the typical operational procedure is to:

\begin{enumerate}
  \item  Obtain, for example, the ERNIE pretrained model
  \item  Fine-tuning it on specific application-relevant data
  \item  More fine-tuning but now in QAT modality, acquiring all the parameters necessary for the subsequent quantization
  \item  Quantize the model
\end{enumerate}

Note that step 2) and 3) can be collapsed. Once reached point 4) it is possible to deploy the obtained model in low-precision inference mode, for example using INT8 as numerical format. \textbf{But with these settings, where could we add the Kurtosis regularizer?} 

\begin{figure}
\centering
\includegraphics[width=1.0 \linewidth, height=9cm]{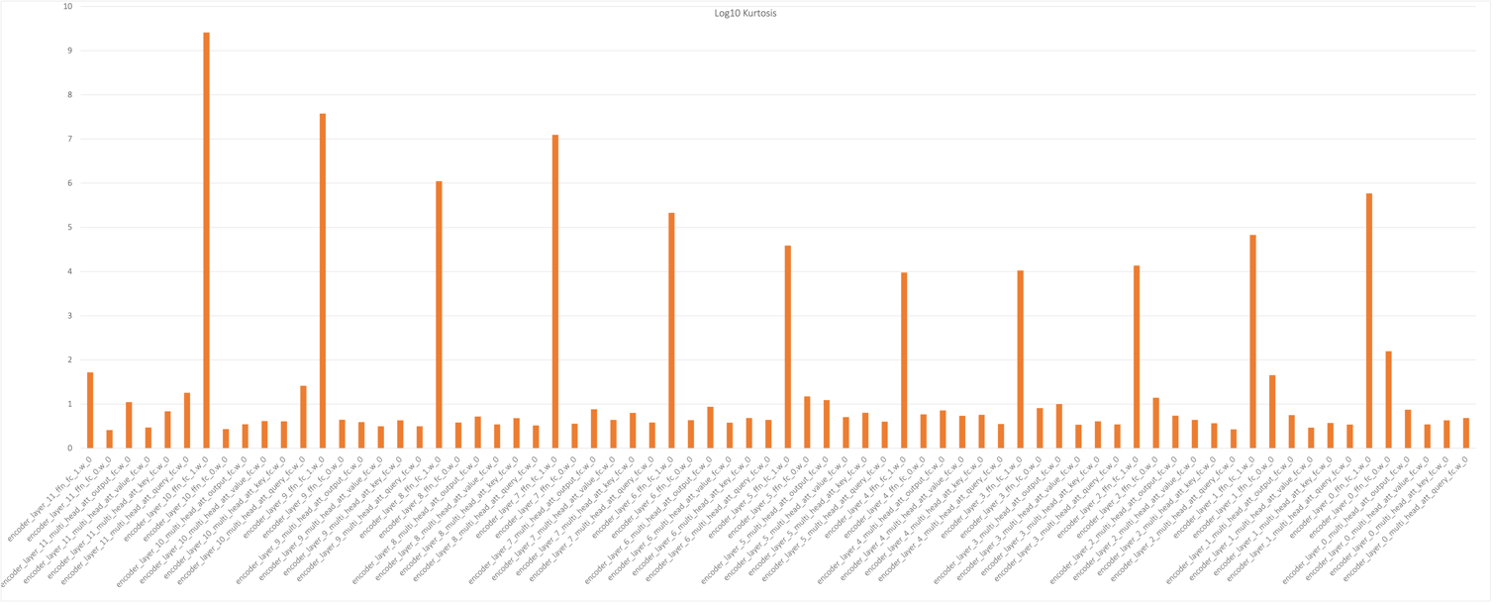} 
\caption{\label{fig:spikes} Kurtosis in log10 scale: it can be seen that some of the tensors (it is FFN\_FC\_1 weight tensor in each block) have Kurtosis of orders of magnitude higher than the others. Note: from left to right we move from the top layer to the bottom layer in ERNIE. There are 12 tensors with Kurtosis above 100 (above 2 in log10 scale): these are the FFN\_FC\_1 weight tensor in each ERNIE block, with ERNIE being composed of 12 blocks}
\vspace{-0.8\baselineskip}
\end{figure}

We tested the options to add Kurtosis Regularizer to step 2) or adding it to step 2) and and 3) when collapsed, gathering better results in the latter case. However, it must be noticed that the naïve application of the Kurtosis regularizer to pre-trained models such ERNIE may not always work out of the box. The reason why this is like so (for example in the ERNIE case), it is connected to the distribution that ERNIE parameters have after the pretraining. Indeed, if we calculate the overall cost component due to the Kurtosis regularizer using the Ernie pretrained parameters we obtain a number that is in the order of 10E10. See fig. \ref{fig:spikes} for a visualization of the Kurtosis values of the pretrained weight tensors in the ERNIE case. This is a huge number and whatever cost function is, it is very unlikely that components of the gradients in such an unbalanced scenario will be appropriately steering the training, and in deeds the training does not proceed in this case. Of course, taking this as black box issue, one could try to find an appropriate lambda (regularizer coefficient) but this is not very practical (we tried!) and does not offer any clarification or understanding of what is going on. And in practice, it does not work at all! Therefore, a bit more of investigation is necessary in order to understand why the Kurtosis behaves like so in the ERNIE case.

\begin{figure}
\centering
\includegraphics[width=1.0\linewidth, height=12cm]{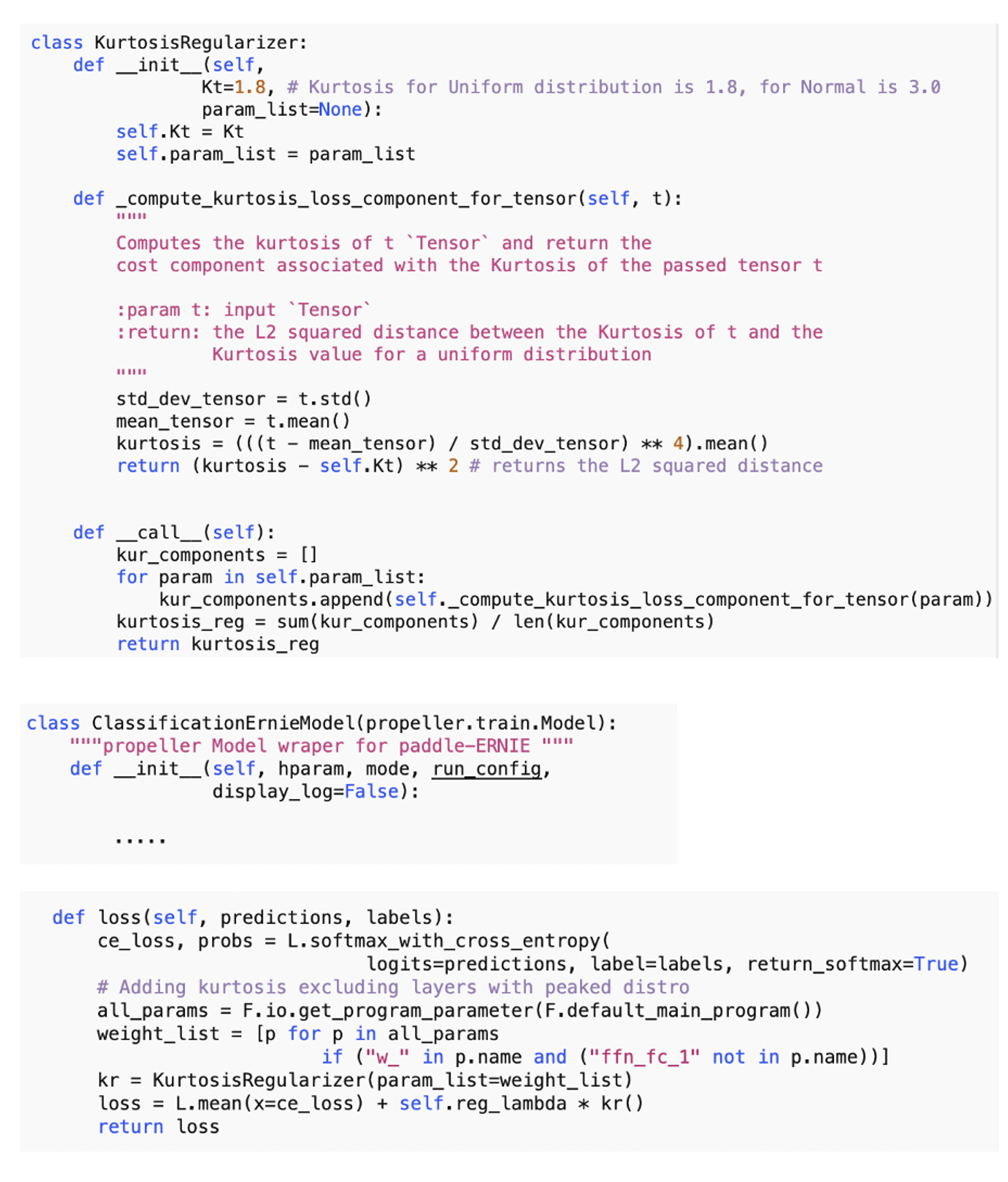} 
\caption{\label{fig:kurpy} Concept of how Kurtosis could be added to a loss function in Paddle, \textbf{showing also how to select a subset of tensors for its computation}.}
\end{figure}

In fig. \ref{fig:kurpy} we show an idea about how Kurtosis could be computed using the PaddlePaddle Deep Learning framework, showing also a possible way to select what tensors to include in the computation, based on the tensors’ name.

The overall Kurtosis applied to ERNIE parameters is the sum the Kurtosis of each parameter tensor used at each layer of the network.
In the context of a model which encompasses many, say M, tensors of parameters, we can then consider the total (or the average) Kurtosis, which is computed as the sum of the Kurtosis for each tensor of parameters.

\begin{eqnarray}
TotalModelKurtosis =  \sum\limits_{j=1}^{M} Kurtosis\left( T_k \right) 
\end{eqnarray}

Looking at each tensor distribution independently, it can be noticed that most of the tensors after the ERNIE pre-training have Kurtosis which is manageable and within reasonable orders of magnitude (say, values are within 1-100) whereas some tensors happen to have Kurtosis in the order of 10E7 or higher.

\begin{figure}
\centering
\includegraphics[width=1.0\linewidth, height=4cm]{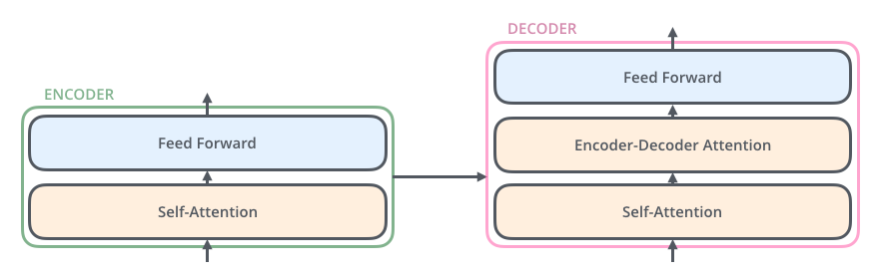} 
\caption{\label{fig:transf_HL} Transformer Encoder block and Decoder block. ERNIE and BERT are composed of only encoder blocks. The picture is taken from https://jalammar.github.io/illustrated-transformer/.}
\end{figure}

We can visualize the distribution of those tensors to better understand what it is going on. As we will see, looking at the list of tensors and their kurtosis, it turns out that there is a regularity in the distribution of the Kurtosis values. 
But let’s not run ahead of ourselves: before delving into the details of the distribution of Kurtosis values, for the sake of clarify it is worth reviewing some concepts about ERNIE and BERT models. They basically consist of a pile (6, 12, 24 or even more) of transformer-encoder blocks, and they are pretrained for different tasks and on a huge amount of data.
Please check the ERNIE paper \cite{ERNIE} for more information.
Each encoder block encompasses a multihead-attention part and a feedforward part, as depicted in fig. \ref{fig:transf_HL}.

\begin{figure}
\centering
\includegraphics[width=.6\linewidth, height=7cm]{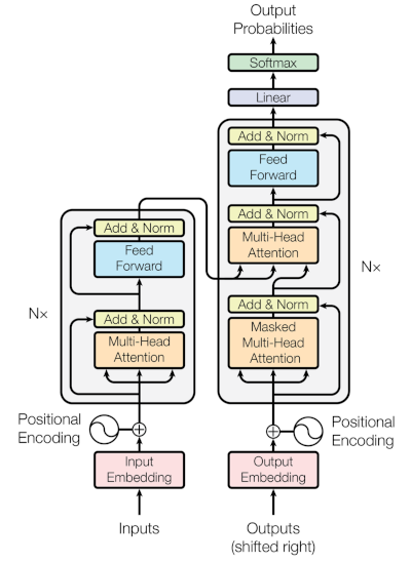} 
\caption{\label{fig:Transformer} Transformer architecture (encoder on the left, decoder on the right) as proposed in \cite{Attention_is_all}.}
\end{figure}

\begin{figure}
\centering
\includegraphics[width=.5\linewidth, height=4cm]{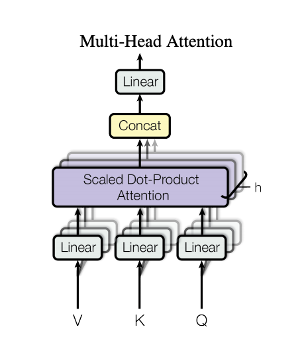} 
\caption{\label{fig:mhead_attention} Multihead attention scheme from \cite{Attention_is_all}.}
\vspace{-0.8\baselineskip}
\end{figure}

\begin{figure}
\includegraphics[width=1.0 \linewidth, height=4cm]{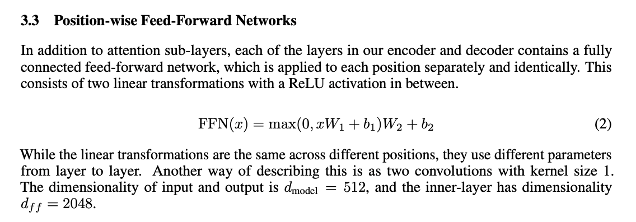} 
\caption{\label{fig:PFFN} Extract from \cite{Attention_is_all}}
\vspace{-0.8\baselineskip}
\end{figure}

Therefore, analyzing the distribution of the tensors of parameters corresponds to analyzing the distribution of the parameters used in the self-attention sub-layer and in the feed-forward sub-layer of each block. More precisely, each self-attention sub-layer is composed of 4 parametrized steps, represented in grey in figure \ref{fig:mhead_attention}, whereas each feed-forward Linear sub-layer is composed of two parametrized steps, as described in the transformer paper already mentioned in figure \ref{fig:PFFN}.

\begin{figure}
\includegraphics[width=1.2 \linewidth, height=18cm]{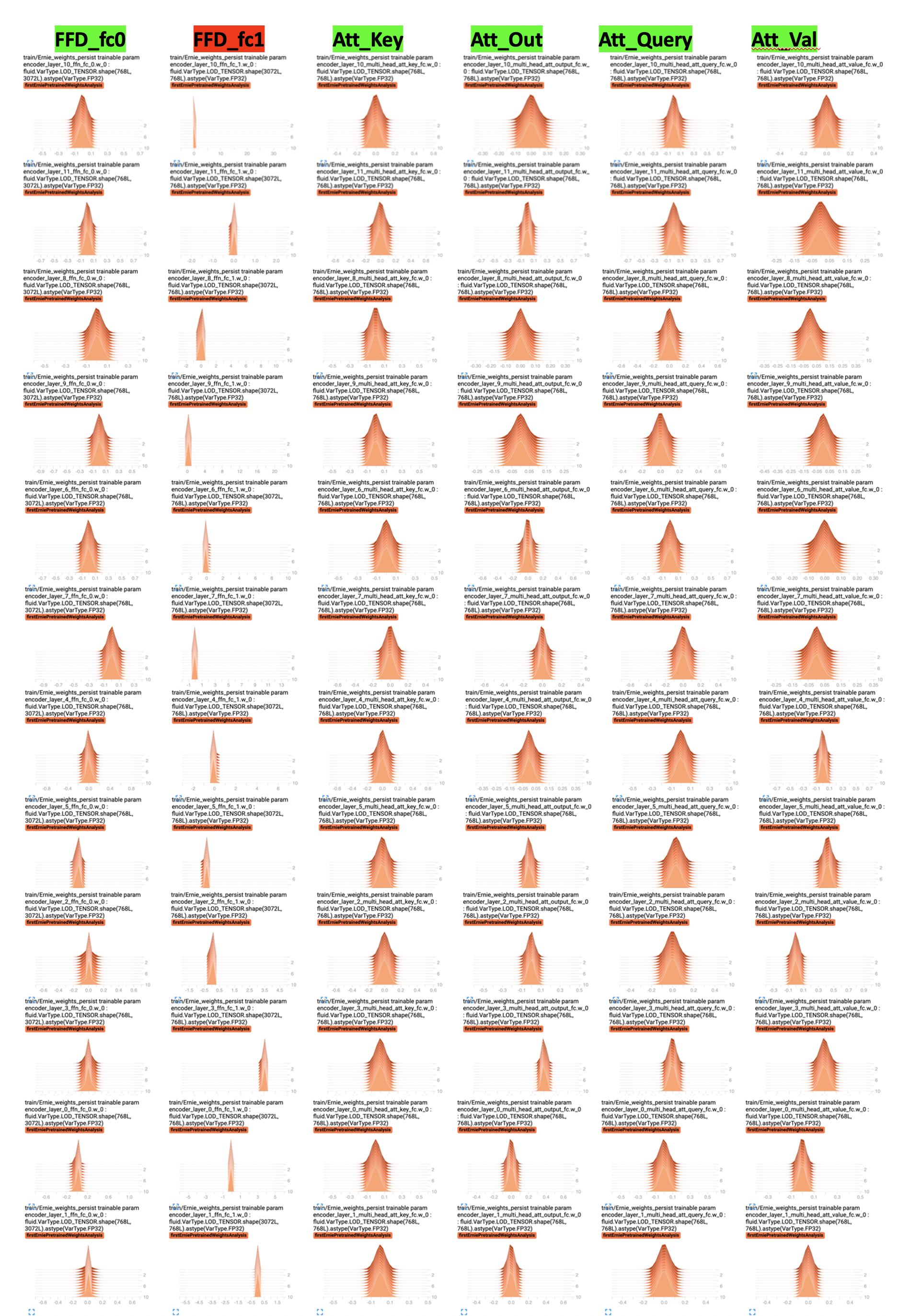} 
\caption{\label{fig:HISTS} Distribution of weighs after ERNIE pretraining}
\vspace{-0.8\baselineskip}
\end{figure}

Therefore, analyzing the distribution of the tensors of parameters corresponds to analyzing the distribution of the parameters used in the self-attention sub-layer and in the feed-forward sub-layer of each block, in other words, of each of the 6 parametrized “steps” in every ERNIE block. A very intuitive way to do it, it is just to visualize it! So let’s have a look at the parameters distributions for all the 12 blocks which we call also layers (as opposed to sub-layers used above) of the Ernie model we used for our experiments (see figure \ref{fig:HISTS}

In figure \ref{fig:HISTS}, from left to right we show the distribution from each encoder block (or layer) of the following:

\begin{enumerate}
  \item	feed-forward sub-layer: parameter tensor for 1st Fully Connected  
  \item	feed-forward sub-layer: parameter tensor for 2nd Fully Connected
  \item	(Multihead) self-attention sub-layer: parameter tensor for Attention Keys
  \item	(Multihead) self-attention sub-layer: parameter tensor for Attention outputs
  \item	(Multihead) self-attention sub-layer: parameter tensor for Attention Queries
  \item	(Multihead) self-attention sub-layer: parameter tensor for Attention Values
\end{enumerate}

\begin{figure}
\includegraphics[width=1. \linewidth, height=5cm]{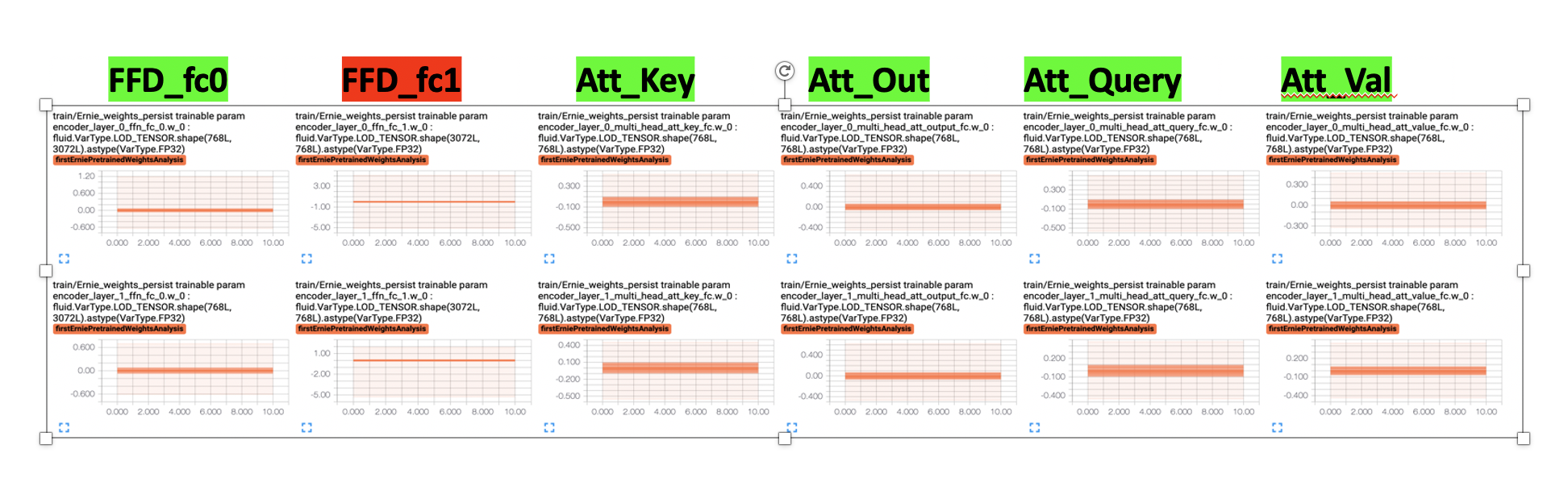} 
\caption{\label{fig:distro} Distribution of weighs after ERNIE pretraining (alternative view)}
\vspace{-0.8\baselineskip}
\end{figure}

As it can be even visually noticed, the Ernie pretraining creates a regular pattern of distributions across blocks and (looking closely) it can be noticed that the parameter tensor for the 2nd Fully Connected in each feed-forward sub-layer has a peculiar distribution, quite peaked and with big very long tails of “outliers”. Outliers are not clearly visible in the histogram chart above, but their presence is the reason why Tensorboard displays a very big interval around the peak in the histogram. 
The existence of wide tails can be better seen using the distribution tab in Tensorboard, to get (for the first block) figure \ref{fig:distro}.

Again, we see that the second Fully Connected in the feed-forward sublayer of the block is extremely peaked with wide tails. 
For such distribution that is very peaked with wide tails, Kurtosis assumes extremely high values due the division by the standard deviation, and therefore it becomes less straightforward to enjoy the effect of Kurtosis regularizer on the fine-tuning step of Ernie.

This can be clearly visualized with a chart that shows how the scale of Kurtosis change as we include critical tensors of parameters in the analysis.

\section{Results and Conclusions}
We now summarize what we have presented so far: we have found that the Ernie pretraining generates a pattern of distributions in the model parameters such that the Fully Connected operations, done as last operation of each ERNIE block, have a peculiar distribution with a peaked core and wide tails. 
This condition is the reason why the overall Kurtosis of the model becomes huge.
Other parameter tensors have Kurtosis that is in the same range of magnitude, making the Kurtosis regularizer easily applicable for them.

The most basic approach to tackle the situation and to make it practical to use Kurtosis regularizer is then to exclude those “critical” tensors from the overall Kurtosis regularizer computation.
This will have the effect to make all the other tensors more “uniformly” distributed at the end of the fine-tuning process, leaving those critical tensors, namely the 2nd Fully Connected tensors in each ERNIE block, adapt during (QAT) fine tuning but without being forced to move towards a more distributed profile.
Please note that we do not investigate here the reason why the pretraining originates this regularity in the tensor distributions: this is a very interesting topic that we leave to future investigations.

One additional point is worth noting: many scenarios are possible, depending on the deployment target we have. Beyond the combination of QAT and Kurtosis, the kind of quantization used, the set of quantizable operations is a “degree of freedom” too.
In this work we take the position that we want to quantize matmul and mul operations, being ERNIE not based on convolutions. 
We report here some numbers relative to what we get in our tests using the Chinese part of the xnli dataset for a task of sentiment classification.

\begin{figure}
\centering
\includegraphics[width=1.0 \linewidth, height=8cm]{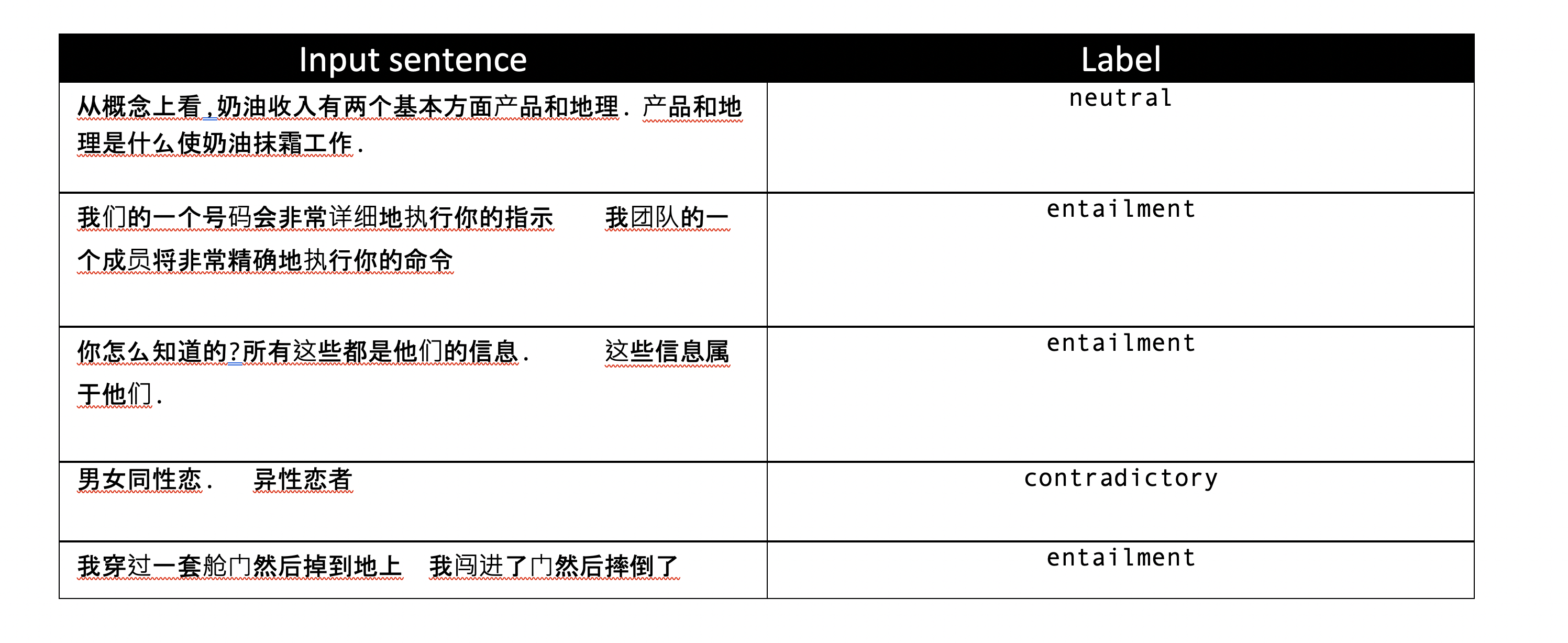} 
\caption{\label{fig:xnli} examples of data entries in xnli dataset}
\vspace{-0.8\baselineskip}
\end{figure}

Here we compare 2 cases:
\begin{enumerate}
  \item	Pre-trained ERNIE followed by fine tuning in QAT with selected Kurtosis, and lambda 0.5, on 60000 batches, with batch-size = 20
  \item	Pre-trained ERNIE followed by fine tuning in QAT with NO Kurtosis on 60000 batches, with batch-size = 20
\end{enumerate}

\begin{figure}
\centering
\includegraphics[width=0.5 \linewidth, height=2cm]{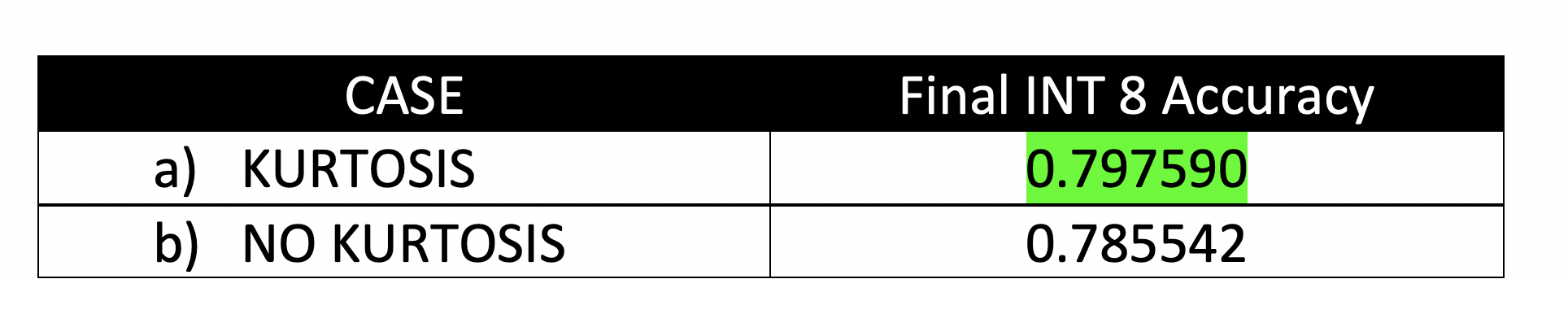} 
\caption{\label{fig:results} effect of Kurtosis regularizer we obtained in our tests}
\vspace{-0.8\baselineskip}
\end{figure}

So the addition of Kurtosis regularizer, applied selectively in the case of ERNIE, has led to an increase to the final INT8 accuracy of 1.2
We are aware of the lack of generality of these results, and their sub-optimality due to the reduced length of the fine-tuning process and the non-optimal batch size used (both due to the limitations of the computational resources we used to for this study) but, in spite of all that, the accuracy gap we obtained in tests is an interesting indicator of the benefits brought by the Kurtosis regularizer also for pre-trained models.

\begin{figure}
\centering
\includegraphics[width=1. \linewidth, height=6cm]{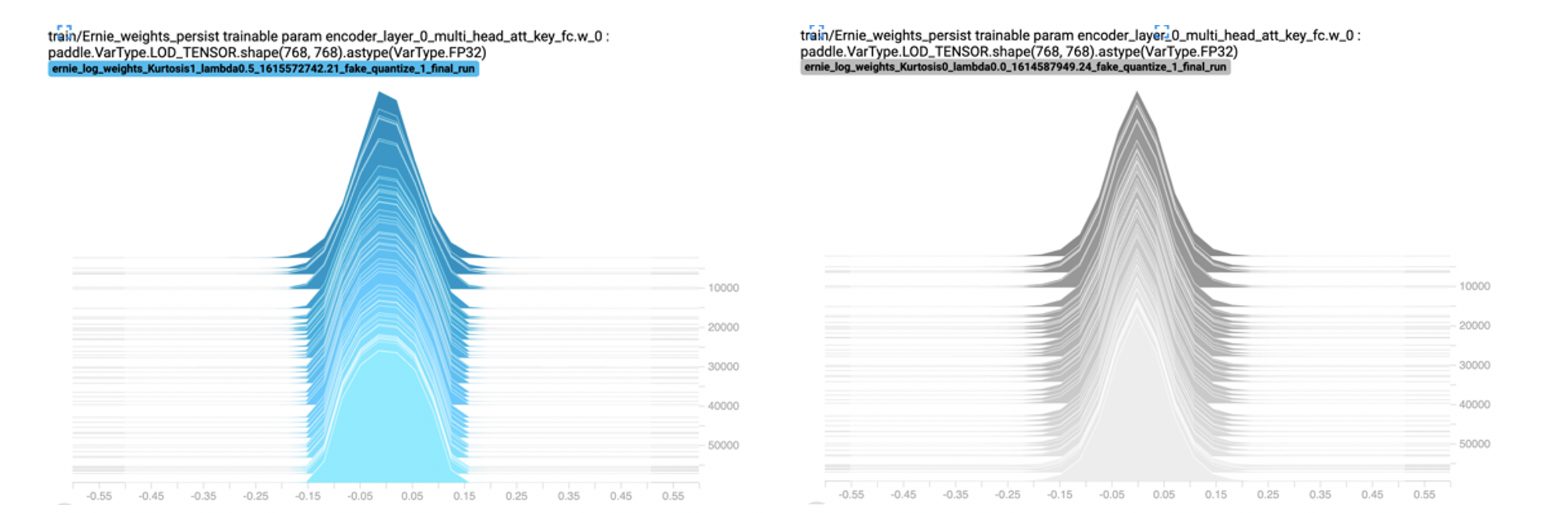} 
\caption{\label{fig:cmp} \textbf{LEFT}: QAT fine tuning of ERNIE with Kurtosis and lambda=0.5 \textbf{RIGHT}: QAT fine tuning of Ernie without kurtosis. It can be noticed how the presence of the Kurtosis regularizer progressively shapes the tensor distribution (in this case we show the “key” part of the Key-Query-Value multi-attention component in layer/block 0 of Ernie. The same effect is present in all tensors that are included in the regularizer}
\vspace{-0.8\baselineskip}
\end{figure}

To effectively visualize the effect of a QAT fine tuning with or without Kurtosis, fig. \ref{fig:cmp} shows the distribution of parameter values changes during the training in the 2 cases.

\end{document}